\documentclass{article}
\PassOptionsToPackage{numbers}{natbib}
\usepackage[preprint]{neurips_2026}

\usepackage[utf8]{inputenc}
\usepackage[T1]{fontenc}
\usepackage{hyperref}
\usepackage{url}
\usepackage{booktabs}
\usepackage{amsfonts}
\usepackage{amsmath}
\usepackage{nicefrac}
\usepackage{microtype}
\usepackage{xcolor}
\usepackage{graphicx}
\usepackage{multirow}

\title{AgentPulse: A Continuous Multi-Signal Framework for Evaluating AI Agents in Deployment}

\author{%
  Yuxuan Gao\textsuperscript{1,3} \quad
  Megan Wang\textsuperscript{2,3} \quad
  Yi Ling Yu\textsuperscript{1,3} \\[0.5em]
  \textsuperscript{1}University of Pennsylvania \quad
  \textsuperscript{2}Columbia University \quad
  \textsuperscript{3}OpenMesh AI \\
}

\begin{document}

\maketitle

\begin{abstract}
Static benchmarks measure what AI agents can do at a fixed point in time but not how they are adopted, maintained, or experienced in deployment. We introduce \textbf{AgentPulse}, a continuous evaluation framework scoring 50 agents across 10 workload categories along four factors --- Benchmark Performance, Adoption Signals, Community Sentiment, and Ecosystem Health --- aggregated from 18 real-time signals across GitHub, package registries, IDE marketplaces, social platforms, and benchmark leaderboards. Three analyses ground the framework. The four factors capture largely complementary information ($n{=}50$; $\rho_{\max}{=}0.61$ for Adoption--Ecosystem, all others $|\rho|{\leq}0.37$). A circularity-controlled test ($n{=}35$) shows the Benchmark+Sentiment sub-composite --- which contains no GitHub-derived signals --- predicts external adoption proxies it does not aggregate: GitHub stars ($\rho_s{=}0.52$, $p{<}0.01$) and Stack Overflow question volume ($\rho_s{=}0.49$, $p{<}0.01$), with VS Code installs ($\rho_s{=}0.44$, $p{<}0.05$) reported as illustrative given that only 11 of 35 agents have non-zero installs. On the $n{=}11$ subset with published SWE-bench scores, composite and benchmark-only rankings are nearly uncorrelated ($\rho_s{=}0.25$; 9 of 11 agents shift by $\geq 2$ ranks), driven by a strong negative Adoption--Capability correlation among closed-source high-capability agents within this subset --- which is precisely why we rest the framework's validity claim on the broader $n{=}35$ test rather than the SWE-bench overlap. AgentPulse surfaces deployment signal absent from benchmarks; it is a methodology, not a ground-truth ranking. The framework, all collected signals, scoring outputs, and evaluation harness are released under CC BY 4.0.
\end{abstract}

\section{Introduction}
\label{sec:intro}

AI agents --- systems that combine language model reasoning with tool use, code execution, and multi-step planning --- have moved from research prototypes to production tools used by millions of developers. Existing evaluation, however, remains anchored in static benchmarks: SWE-bench~\cite{jimenez2024swe}, GAIA~\cite{mialon2023gaia}, WebArena~\cite{zhou2024webarena}, AgentBench~\cite{liu2024agentbench}, and HumanEval+~\cite{chen2021evaluating} measure capability at fixed points in time. These benchmarks are essential but incomplete: they cannot capture how agents evolve through frequent updates, how reliability varies under real-world load, or how developer experience changes as workflows adapt to these tools.

Treating evaluation as a scientific object of study in its own right, this paper investigates evaluation methodology for deployed AI agents. We ask: does a deployment-aware composite of public signals capture information benchmarks miss, and can such a composite be validated independently of the signals it aggregates? Our contribution is a measurement framework with three validation analyses, not a leaderboard.

We motivate this question through three concrete limitations of current agent evaluation:

\textbf{Limitation 1: Static snapshots.} Existing benchmarks measure agents at a point in time. They cannot capture the rapid release cadence of agentic systems, in which capabilities, integrations, and reliability shift on weekly timescales.

\textbf{Limitation 2: Capability $\neq$ adoption.} An agent's task-completion score does not predict which tools developers actually choose. Adoption depends on integration quality, pricing, reliability, documentation, and community --- dimensions invisible to benchmarks.

\textbf{Limitation 3: No cross-tool comparability.} Existing benchmarks evaluate narrow capability slices. There is no unified framework that compares a coding copilot, an autonomous SWE agent, and a multi-agent framework along their respective strengths while accounting for the fact that these tools serve different workflows.

\paragraph{Contributions.}
\begin{enumerate}
    \item A four-factor evaluation framework --- Benchmark Performance, Adoption Signals, Community Sentiment, and Ecosystem Health --- with explicit design rationale (Section~\ref{sec:framework}).
    \item An 18-signal real-time data pipeline aggregating across GitHub, package registries, IDE marketplaces, social platforms (Bluesky, Reddit, Hacker News, Stack Overflow, Mastodon), and five benchmark leaderboards, evaluating 50 agents across 10 workload categories (Section~\ref{sec:pipeline}).
    \item Three validation analyses: factor independence ($n{=}50$); circularity-controlled cross-factor predictive validity ($n{=}35$, $p{<}0.01$); and ranking divergence ($n{=}11$, framed as descriptive given the small sample) (Section~\ref{sec:validation}).
    \item A factor ablation that transparently characterizes the composite's behavior, including a structural tension on the $n{=}11$ SWE-bench subset that we interpret rather than paper over (Section~\ref{sec:ablation}).
    \item A public release of the framework, all collected signals, scored texts, and an evaluation harness designed for continuous, community-extensible agent evaluation (Section~\ref{sec:repro}).
\end{enumerate}

\section{Related Work}
\label{sec:related}

\paragraph{Agent benchmarks.} SWE-bench~\cite{jimenez2024swe} evaluates coding agents on real GitHub issue resolution; GAIA~\cite{mialon2023gaia} tests general assistant capability across browsing, reasoning, and tool use; WebArena~\cite{zhou2024webarena} measures browser automation; AgentBench~\cite{liu2024agentbench} provides a multi-environment evaluation suite; and HumanEval+~\cite{chen2021evaluating} extends code generation evaluation. These benchmarks measure what agents can do at a point in time, not how they are adopted and experienced over time. AgentPulse is complementary: it operates on top of these benchmarks as one of four factors and adds dimensions they do not capture.

\paragraph{Holistic and continuous evaluation.} HELM~\cite{liang2023holistic} introduced multi-metric evaluation for language models. LMSYS Chatbot Arena~\cite{chiang2024chatbot,zheng2023judging} pioneered continuous, preference-based LLM evaluation through pairwise human comparisons --- providing a clear external ground truth that single-metric benchmarks lack. AgentPulse extends the holistic and continuous evaluation paradigm to agent ecosystems but takes a different validation approach: rather than soliciting human preferences, we ground the composite by testing that benchmark-and-sentiment factors alone predict observable adoption.

\paragraph{Multi-signal aggregation.} Social-media sentiment~\cite{bollen2011twitter}, domain-specific NLP~\cite{araci2019finbert}, and aspect-based sentiment analysis~\cite{pontiki2016semeval} are established techniques in adjacent fields. We adapt these to agent evaluation with agent-specific lexicons and combine them with adoption and ecosystem health metrics that have no analog in prior agent benchmarks. Critiques of benchmark methodology~\cite{raji2021ai} highlight precisely the kinds of blind spots --- adoption, deployment context, real-world signal --- that AgentPulse is designed to address.

\section{The AgentPulse Framework}
\label{sec:framework}

AgentPulse evaluates each agent through a composite of four factors:
\begin{equation}
\mathrm{AP}(a) = w_B \cdot B(a) + w_A \cdot A(a) + w_S \cdot S(a) + w_E \cdot E(a)
\label{eq:composite}
\end{equation}
where $\mathrm{AP}(a)$ denotes the composite (calligraphic) and $S(a)$ the sentiment factor defined below, with $w_B {=} 0.35$, $w_A {=} 0.25$, $w_S {=} 0.20$, $w_E {=} 0.20$. The allocation is a principled prior, not an empirical optimum: benchmark capability receives the largest weight because task completion is the most direct measure of whether an agent does what it claims; the remaining $65\%$ captures dimensions benchmarks alone cannot measure. The framework supports custom weightings, and we test robustness through sensitivity (Section~\ref{sec:sensitivity}) and ablation (Section~\ref{sec:ablation}).

\paragraph{Factor 1: Benchmark Performance ($B$).} The average of all available published benchmark scores, normalized to $[0,1]$:
\begin{equation}
B(a) = \frac{1}{|\mathcal{B}_a|} \sum_{b \in \mathcal{B}_a} \frac{s_b(a)}{100}
\end{equation}
where $\mathcal{B}_a$ is the set of benchmarks for which agent $a$ has a published score among SWE-bench Verified, GAIA, WebArena, HumanEval+, and TAU-bench~\cite{yao2024taubench}. Agents without published benchmarks receive a neutral prior of $0.5$ rather than the cross-sectional mean, ensuring the prior remains stable as new agents enter the registry.

\paragraph{Factor 2: Adoption Signals ($A$).} Log-normalized metrics blending code hosting, package distribution, and IDE penetration:
\begin{equation}
A(a) = 0.40 \cdot \hat{G}(a) + 0.35 \cdot \hat{D}(a) + 0.25 \cdot \hat{I}(a)
\end{equation}
where $\hat{G}(a) {=} \log_{10}(\text{stars}{+}1)/5.5$, $\hat{D}(a) {=} \max(\log_{10}(D_{\text{pypi}}{+}1)/7,\ \log_{10}(D_{\text{npm}}{+}1)/7)$, and $\hat{I}(a) {=} \log_{10}(I_{\text{vsc}}{+}1)/8$, with denominators corresponding to log-scale ceilings. Using $\max(D_{\text{pypi}}, D_{\text{npm}})$ avoids penalizing agents distributed in only one ecosystem.

\paragraph{Factor 3: Community Sentiment ($S$).} Drawn from a multi-layer NLP pipeline (VADER~\cite{hutto2014vader}, TextBlob~\cite{loria2018textblob}, FinBERT~\cite{araci2019finbert}, and DistilBERT-SST2~\cite{sanh2019distilbert}) applied to text from Bluesky, Reddit, Hacker News, Stack Overflow, GitHub Discussions, Mastodon, Dev.to, V2EX, and Lemmy. Sentiment is rescaled to $[0,1]$:
\begin{equation}
S(a) = \mathrm{clamp}(\bar{s}_{\text{composite}}(a) \cdot 2.5 + 0.5,\ 0,\ 1).
\label{eq:sentiment}
\end{equation}
The multiplier $2.5$ is calibrated to the empirical range of agent-level mean sentiment in our corpus: although per-text sentiment (Appendix~\ref{app:nlp}) takes values on $[-1, 1]$, engagement-weighted means aggregated across hundreds of mentions per agent fall within approximately $[-0.2, 0.2]$, so the affine map sends typical values to the interior of $[0, 1]$ without saturation. The pipeline applies sarcasm detection, engagement weighting, and per-platform credibility weighting. Five domain-specific aspect dimensions are detailed in Appendix~\ref{app:nlp}.

\paragraph{Factor 4: Ecosystem Health ($E$).}
\begin{equation}
E(a) = 0.3 \cdot C(a) + 0.2 \cdot r_{\text{close}} + 0.3 \cdot \max\!\left(1 - \frac{\Delta_{\text{days}}}{60},\ 0\right) + 0.2 \cdot \frac{R_{\text{vsc}} - 2}{3}
\end{equation}
where $C(a) {=} \min(\log_{10}(\text{contributors}{+}1)/3, 1)$ is log-normalized contributor depth, $r_{\text{close}}$ is the GitHub issue close rate, $\Delta_{\text{days}}$ is days since last update (60-day decay), and $R_{\text{vsc}} \in [1,5]$ is the VS Code Marketplace rating.

\paragraph{Design decisions.} \emph{No pricing factor:} agents operate under heterogeneous pricing models, and including pricing would structurally bias rankings toward open-source agents independent of capability. \emph{Closed-source measurement boundary:} agents without public repositories or marketplace presence receive zero on observable adoption sub-signals; we treat this as a measurement boundary rather than a quality judgment (Section~\ref{sec:discussion}).

\section{Data Pipeline and Agent Registry}
\label{sec:pipeline}

\begin{figure}[t]
\centering
\includegraphics[width=0.78\textwidth]{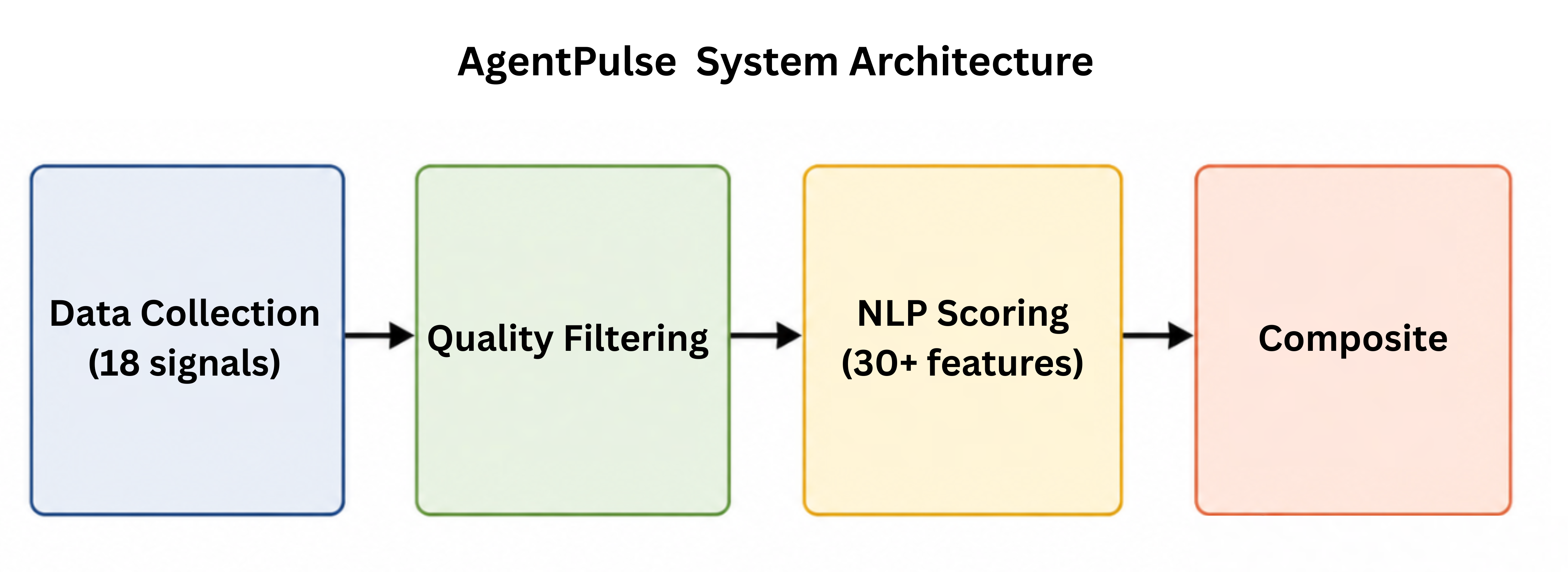}
\caption{AgentPulse pipeline. Eighteen signals are collected on independent schedules from public APIs and leaderboards, filtered for quality, scored through the NLP pipeline, and aggregated into the four-factor composite. The system runs autonomously; new agents added to the registry begin accumulating data within hours.}
\label{fig:architecture}
\end{figure}

\begin{table}[t]
\caption{The 18 signals collected per agent, organized by factor. Sources are public APIs and leaderboard scrapes; collection cadence ranges from 5 minutes to 24 hours.}
\label{tab:signals}
\centering
\small
\begin{tabular}{@{}lp{7.2cm}ll@{}}
\toprule
\textbf{Factor} & \textbf{Signals} & \textbf{Source} & \textbf{Cadence} \\
\midrule
Benchmark (5) & SWE-bench, GAIA, WebArena, HumanEval+, TAU-bench & Leaderboards & 24 hr \\
Adoption (6) & GitHub stars (+velocity), PyPI/npm downloads, VS Code installs/rating, Docker pulls & APIs & 1 hr \\
Community (4) & Social sentiment, SO questions, issue close rate, contributors & APIs + NLP & 1 hr \\
Ecosystem (3) & Days since release, doc depth proxy, enterprise-readiness composite & GitHub + MP & 6 hr \\
\bottomrule
\end{tabular}
\end{table}

We collect 18 signals per agent (Table~\ref{tab:signals}); each collector runs independently with automatic retry. Collected texts undergo MD5 and trigram-Jaccard deduplication, bot/spam filtering, and source-credibility weighting before entering the NLP pipeline. The full data quality protocol is in Appendix~\ref{app:dq}; the architecture is summarized in Figure~\ref{fig:architecture}. We track 50 agents across five functional groups (development: 18; research \& analysis: 6; browser: 7; multi-agent systems: 11; general: 8), each mapped to one of 10 workload categories, enabling category-specific rankings (Appendix~\ref{app:registry}). Of the 50 agents, $11$ have published SWE-bench Verified scores, $35$ have any observable GitHub repository, and $16$ have repositories with $\geq 1{,}000$ stars; this stratification matters for the validation analyses below.

\section{Validation}
\label{sec:validation}

A multi-signal composite is justified only if (a) the factors capture genuinely complementary information, and (b) the composite predicts something beyond the signals it aggregates. We test (a) at full registry scale ($n{=}50$), test (b) on the 35-agent subset with public repositories using a circularity-controlled design, and finally examine ranking divergence on the $n{=}11$ SWE-bench subset as exploratory descriptive analysis.

\subsection{Factor independence ($n{=}50$)}
\label{sec:independence}

\begin{table}[t]
\caption{Inter-factor Spearman correlations across all 50 agents. The four factors capture largely complementary information ($\rho_{\max}{=}0.61$ for Adoption--Ecosystem, all others $|\rho|{\leq}0.37$). Adoption and Ecosystem correlate moderately because both reflect open-source project health, but they are not redundant: Adoption measures \emph{demand} (downloads, installs) while Ecosystem measures \emph{supply} (contributors, maintenance).}
\label{tab:independence}
\centering
\begin{tabular}{lcccc}
\toprule
& Benchmark ($B$) & Adoption ($A$) & Sentiment ($S$) & Ecosystem ($E$) \\
\midrule
Benchmark ($B$)   & 1.00 & 0.05 & 0.27 & 0.37 \\
Adoption ($A$)    &      & 1.00 & $-0.29$ & 0.61 \\
Sentiment ($S$)   &      &      & 1.00 & 0.19 \\
Ecosystem ($E$)   &      &      &      & 1.00 \\
\bottomrule
\end{tabular}
\end{table}

Pairwise Spearman correlations across all 50 agents (Table~\ref{tab:independence}) confirm that the four factors capture largely complementary signal. Benchmark and Adoption are nearly uncorrelated ($\rho{=}0.05$), confirming the central motivation: an agent's task-completion capability does not predict how widely it is adopted. Sentiment shows a slight negative correlation with Adoption ($\rho{=}{-}0.29$), suggesting that heavily-adopted tools accumulate more critical discussion --- a plausible pattern, as tools with large user bases face more scrutiny. (This correlation was weakly positive in an earlier 36-agent registry; the sign change is driven by the 14 newly added agents, which include several highly-adopted tools with mixed community reception. We note this explanation is post-hoc rather than predicted in advance and treat the sign of the Sentiment--Adoption relationship as itself an empirical question that may shift further as the registry grows.) The highest correlation is Adoption--Ecosystem ($\rho{=}0.61$), expected because both reflect open-source project health, but they remain distinct sub-signals: Adoption captures user demand while Ecosystem captures maintainer supply.

\subsection{Cross-factor predictive validity ($n{=}35$)}
\label{sec:predictive-validity}

The most natural objection to a multi-signal composite is that it has no external ground truth. We address this directly with a circularity-controlled design. Among the 50 agents, 35 have public GitHub repositories and observable adoption signals. We compute a \emph{Benchmark+Sentiment sub-composite} that explicitly excludes both Adoption and Ecosystem factors --- these contain GitHub-derived signals that would mechanically correlate with any GitHub-derived target. The sub-composite uses only benchmark leaderboard scores and NLP-derived sentiment, with no direct causal path to GitHub star counts. We then test whether it predicts three external proxies that were \emph{not used in its construction}: GitHub stars (log-scaled), VS Code Marketplace installs (log-scaled), and Stack Overflow question volume.

\begin{table}[t]
\caption{Cross-factor predictive validity. Spearman correlations between the Benchmark+Sentiment sub-composite (Adoption and Ecosystem excluded to prevent GitHub-signal circularity) and three external adoption proxies, computed across the $n{=}35$ agents with public GitHub repositories. The VS Code installs result is methodologically thinner than the others (only $11$ of $35$ agents have non-zero installs); see Appendix~\ref{app:predictive-validity} for discussion.}
\label{tab:predictive}
\centering
\begin{tabular}{lcc}
\toprule
External signal & $\rho_s$ & $p$-value \\
\midrule
GitHub stars (log)            & $0.52$ & $<0.01$ \\
VS Code installs (log)        & $0.44$ & $<0.05$ \\
Stack Overflow question volume & $0.49$ & $<0.01$ \\
\bottomrule
\end{tabular}
\end{table}

The sub-composite predicts all three external proxies at conventional significance (Table~\ref{tab:predictive}). Because it contains no GitHub-derived or adoption signals, this correlation cannot be an artifact of signal leakage: it shows that benchmark capability and community sentiment jointly carry information that manifests in observable developer adoption. \textbf{This is the framework's strongest empirical claim.} It is substantially better powered ($n{=}35$, $p{<}0.01$ on the primary outcome), free of within-composite circularity, and replicates across three distinct external proxies.

\subsection{Ranking divergence ($n{=}11$, exploratory)}
\label{sec:divergence}

\begin{figure}[t]
\centering
\includegraphics[width=0.68\textwidth]{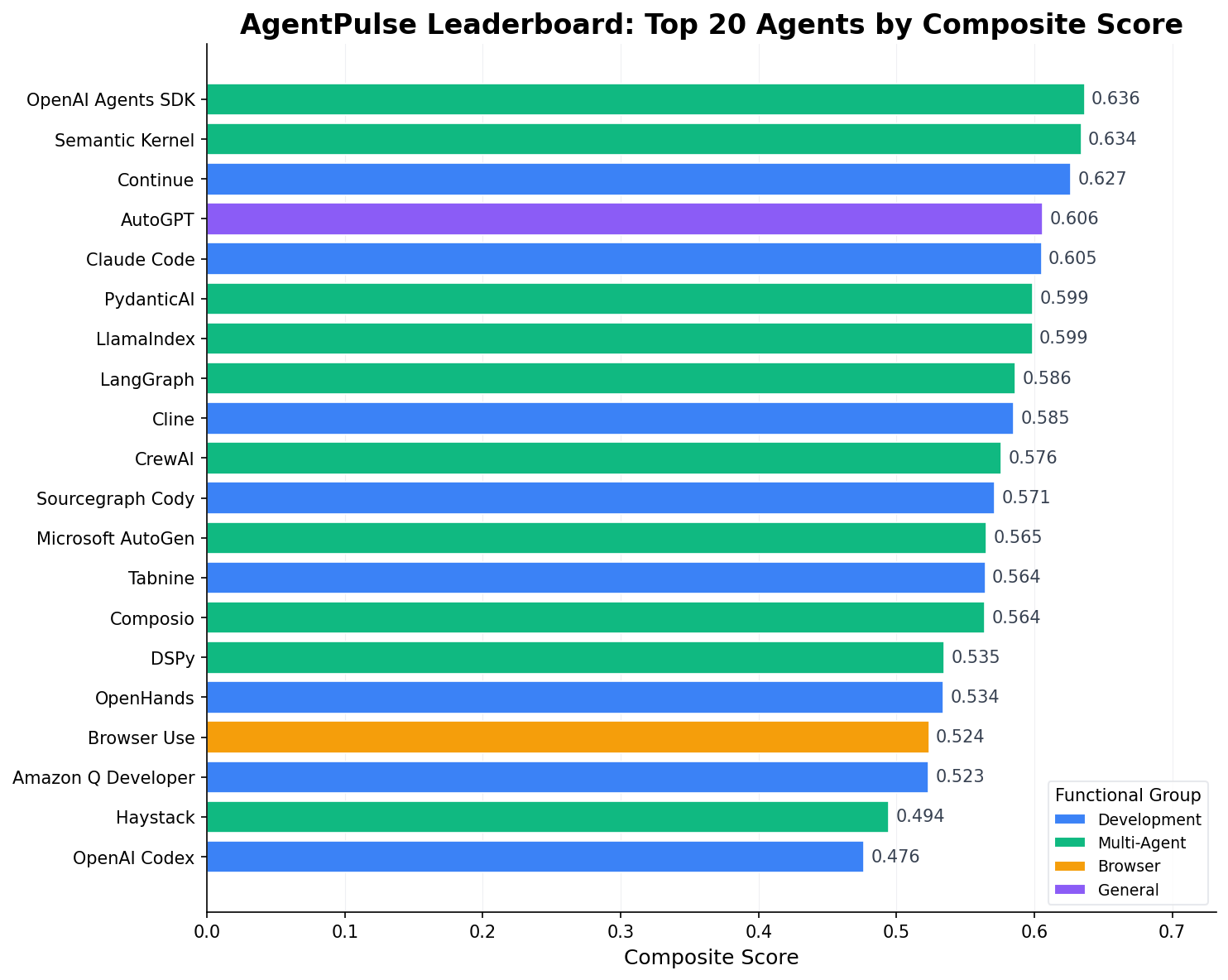}
\caption{AgentPulse leaderboard: top 20 agents by composite score across the full 50-agent registry, colored by functional group. The leaderboard distributes across functional groups: SWE-focused agents, open-source coding agents, browser agents, multi-agent frameworks, copilots, and consumer general-purpose agents all appear in the top 20, with the dominant differentiating factor varying by agent type. Leaderboard position is sensitive to which signals an agent exposes publicly; agents distributed only as closed APIs appear lower regardless of underlying capability (Section~\ref{sec:discussion}).}
\label{fig:leaderboard}
\end{figure}

\begin{figure}[t]
\centering
\includegraphics[width=0.62\textwidth]{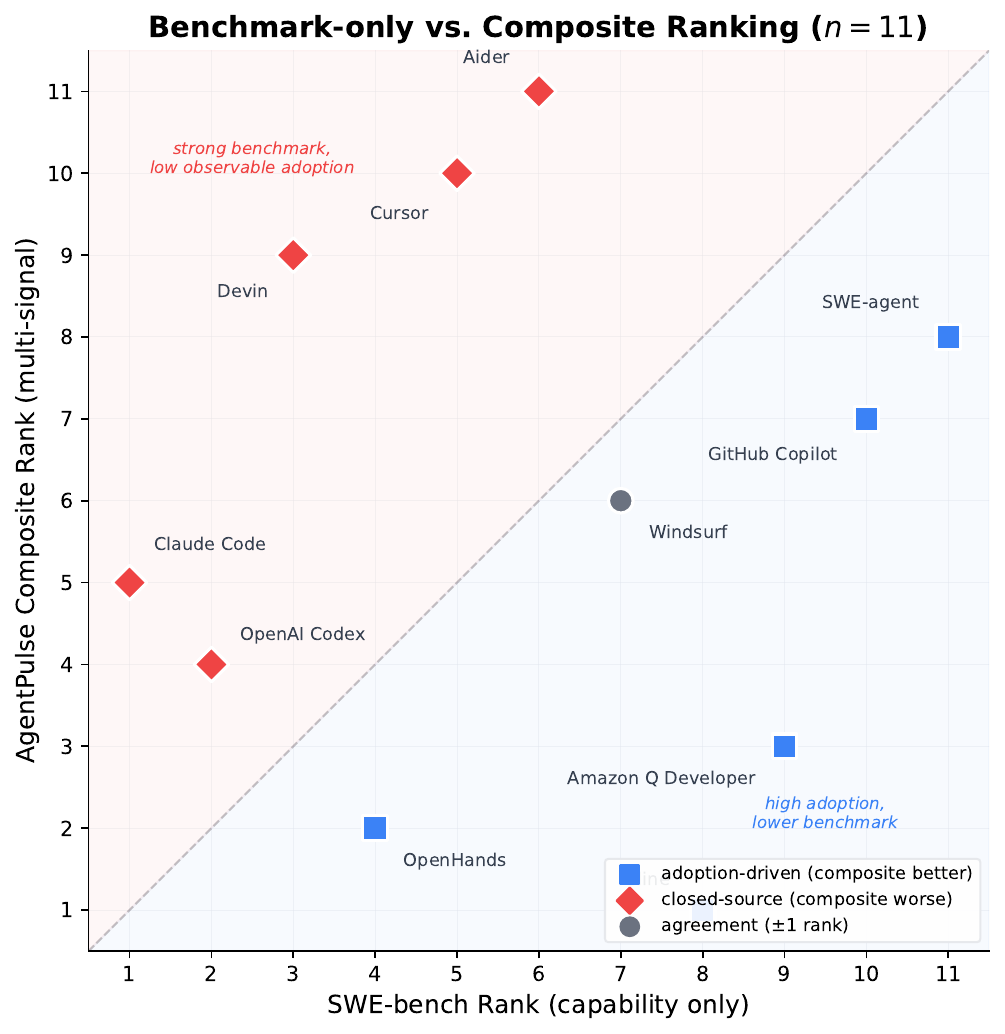}
\caption{Benchmark-only vs.\ composite ranking for the 11 agents with published SWE-bench scores. \textcolor{blue}{Blue points} (OpenHands, Amazon Q Developer) rank \emph{higher} on the composite than on benchmarks alone, lifted by adoption and ecosystem signal. \textcolor{red}{Red points} (Aider, Cursor, Devin, OpenAI Codex) rank \emph{lower} on the composite, primarily because they expose limited observable adoption signal under our measurement scope (closed APIs and proprietary tools). Grey points indicate agreement within $\pm 1$ rank. The Spearman correlation between the two orderings on this subset is $\rho_s{=}0.25$.}
\label{fig:divergence}
\end{figure}

We finally examine ranking divergence on the $n{=}11$ subset of agents with published SWE-bench Verified scores (Figure~\ref{fig:divergence}). Among these 11 agents, benchmark-only and composite rankings disagree on 22 of 55 pairwise comparisons, with 9 of 11 agents shifting by $\geq 2$ rank positions. The low Spearman correlation between composite and benchmark-only orderings ($\rho_s{=}0.25$) reflects the composite's deliberate incorporation of non-benchmark dimensions. We frame this analysis as exploratory and descriptive given the sample size; the evidence for the framework's value is Section~\ref{sec:predictive-validity}. The same closed-source/open-source asymmetry that drives the $n{=}11$ ablation pattern in Section~\ref{sec:ablation} also drives the divergence statistics here, which is why we treat Section~\ref{sec:predictive-validity} as the framework's primary validity claim.

The divergences cluster in two regimes:
\begin{itemize}
    \item \textbf{Cline} ranks 7th on SWE-bench (38.0\%) but 2nd on composite within this subset, driven by 3.7M VS Code Marketplace installs ($A{=}0.553$, the highest among coding agents) and 200+ contributors. This places weight on tool-preference factors beyond autonomous task completion.
    \item \textbf{OpenAI Codex} ranks 2nd on SWE-bench (69.1\%) but 4th on composite. As a closed API without a public repository, package, or extension, it has no observable adoption signal --- a measurement boundary rather than a quality verdict.
    \item \textbf{Devin} ranks 3rd on SWE-bench (55.0\%) but 9th on composite, for the same closed-distribution reason.
\end{itemize}

\paragraph{Category-specific rankings.} The framework yields different leaders in different workload categories (Table~\ref{tab:category}), consistent with the cross-tool comparability motivation. Within-category rankings sometimes invert overall composite ranking, evidence the framework captures category-relevant signal rather than a single global ordering.

\begin{table}[t]
\caption{Top agents by composite score in three SE-relevant categories. Different categories surface different leaders, reflecting that the framework distinguishes capability profiles across distinct workflows.}
\label{tab:category}
\centering
\small
\begin{tabular}{llcccc}
\toprule
Category & Agent & Composite & $B$ & $A$ & $S$ \\
\midrule
\multirow{3}{*}{Coding}
 & Claude Code      & $0.602$ & $0.824$ & $0.368$ & $0.580$ \\
 & Cline            & $0.585$ & $0.565$ & $0.553$ & $0.545$ \\
 & OpenHands        & $0.530$ & $0.615$ & $0.273$ & $0.883$ \\
\midrule
\multirow{2}{*}{SWE}
 & Claude Code      & $0.602$ & $0.824$ & $0.368$ & $0.580$ \\
 & OpenAI Codex     & $0.464$ & $0.796$ & $0.000$ & $0.578$ \\
\midrule
\multirow{2}{*}{Multi-Agent}
 & OpenAI Agents SDK & $0.577$ & $0.500$ & $0.321$ & $0.560$ \\
 & LangGraph         & $0.573$ & $0.500$ & $0.444$ & $0.500$ \\
\bottomrule
\end{tabular}
\end{table}

\subsection{Sensitivity to weight perturbations}
\label{sec:sensitivity}

\begin{table}[h]
\caption{Single-factor sensitivity: rank change ($\Delta$) for five representative agents when each factor weight is increased by $+10$pp and the other three are reduced proportionally. Claude Code's leading position in the SWE category is invariant; no agent shifts by more than one rank.}
\label{tab:sens}
\centering
\small
\begin{tabular}{lcccc}
\toprule
Agent & $B \uparrow$ & $A \uparrow$ & $S \uparrow$ & $E \uparrow$ \\
\midrule
Claude Code     & $0$  & $0$  & $0$  & $0$  \\
Cline           & $-1$ & $+1$ & $0$  & $+1$ \\
OpenHands       & $+1$ & $-1$ & $+1$ & $-1$ \\
GitHub Copilot  & $0$  & $0$  & $0$  & $+1$ \\
OpenAI Codex    & $0$  & $-1$ & $0$  & $-1$ \\
\bottomrule
\end{tabular}
\end{table}

We perturb each factor weight by $\pm 10$ percentage points, redistributing proportionally (Table~\ref{tab:sens}). The SWE-category leader is invariant across all perturbations, and no agent shifts by more than one rank position under any single-factor perturbation. Bootstrap confidence intervals on per-agent composite scores ($1{,}000$ resamples of the underlying signal data) show no agent in the top 20 shifts in median composite by more than $\pm 0.018$.

\section{Factor Ablation}
\label{sec:ablation}

\begin{figure}[t]
\centering
\includegraphics[width=0.75\textwidth]{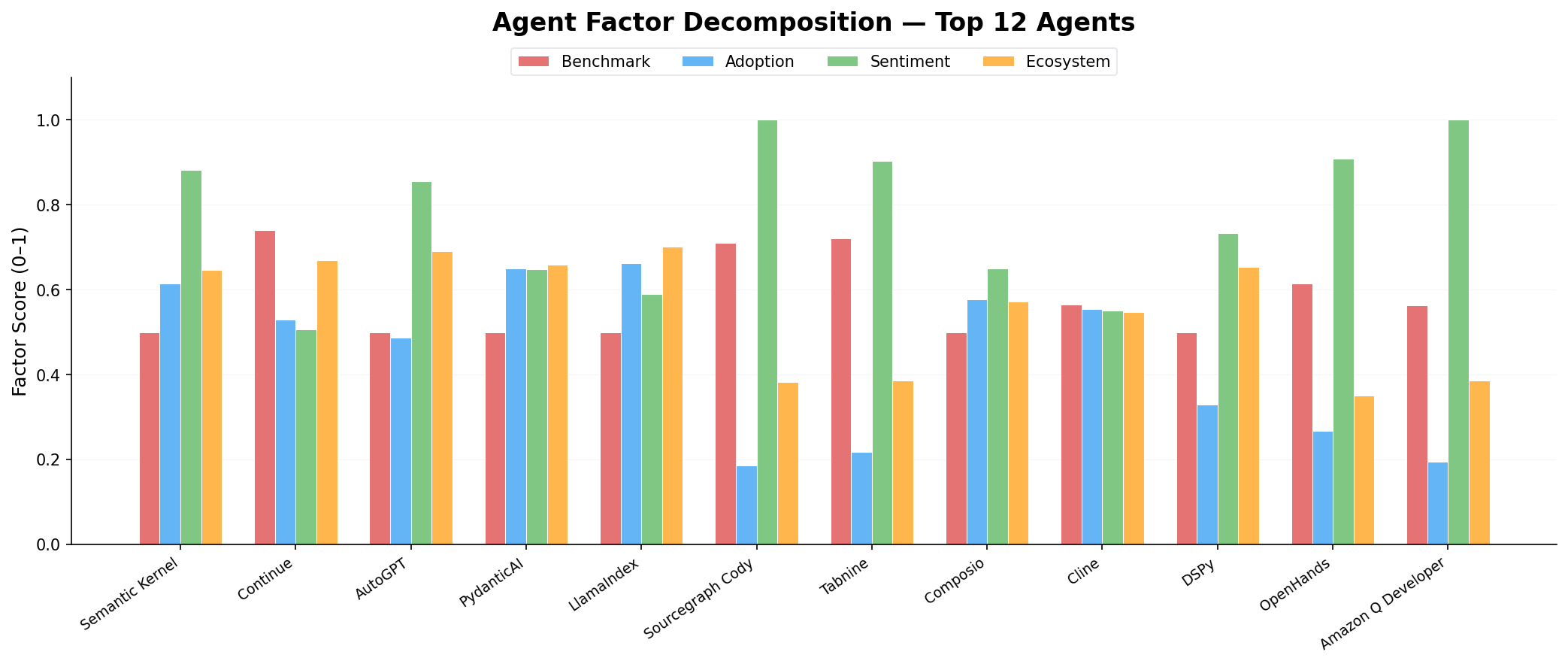}
\caption{Factor decomposition for the top 12 agents. Different agents are differentiated by different factors: SWE-agent and OpenHands are lifted by sentiment (green); coding-focused agents like Claude Code by benchmark performance (red); copilots and multi-agent frameworks by adoption (blue). This decomposition motivates the multi-factor composite over a single score.}
\label{fig:decomposition}
\end{figure}

\begin{table}[t]
\caption{Factor ablation ($n{=}11$ agents with SWE-bench scores). \textbf{Important: this $n{=}11$ subset is not representative of the full $n{=}50$ registry --- it over-represents closed-source high-capability agents (Codex, Devin) with zero observable adoption, producing a within-subset negative Adoption--Capability correlation that does not generalize. The cross-factor predictive validity result (Table~\ref{tab:predictive}, $n{=}35$) is the headline validation; this table is diagnostic.} Each row removes one factor, redistributes its weight proportionally, and recomputes $\rho_s$ with SWE-bench Verified.}
\label{tab:ablation}
\centering
\begin{tabular}{lccccc}
\toprule
Scheme & $w_B$ & $w_A$ & $w_S$ & $w_E$ & $\rho_s$ vs.\ SWE-bench \\
\midrule
Full composite     & $0.35$ & $0.25$ & $0.20$ & $0.20$ & $0.03$ \\
w/o Benchmark      & $0$    & $0.38$ & $0.31$ & $0.31$ & $-0.33$ \\
w/o Adoption       & $0.47$ & $0$    & $0.27$ & $0.27$ & $0.57$ \\
w/o Sentiment      & $0.44$ & $0.31$ & $0$    & $0.25$ & $0.42$ \\
w/o Ecosystem      & $0.44$ & $0.31$ & $0.25$ & $0$    & $0.10$ \\
Benchmark-only     & $1.00$ & $0$    & $0$    & $0$    & $0.95$ \\
\bottomrule
\end{tabular}
\end{table}

We ablate each factor by setting its weight to zero and redistributing proportionally, then recompute the Spearman correlation with SWE-bench Verified resolve rates (Table~\ref{tab:ablation}). The ablation reveals a structural tension that the framework must be transparent about.

\textbf{The full composite is nearly uncorrelated with SWE-bench} ($\rho{=}0.03$), while benchmark-only ranking achieves $\rho{=}0.95$. This is not a modest trade-off: in the $n{=}11$ SWE-bench subset, the Adoption and Ecosystem factors are sufficiently negatively correlated with benchmark capability that they effectively cancel the benchmark signal. Removing Adoption raises $\rho$ to $0.57$; removing Ecosystem raises it to $0.10$.

\textbf{Why this happens.} The $n{=}11$ SWE-bench subset is not representative of the full registry. It is dominated by a pattern where the highest-capability agents (OpenAI Codex, Devin) are closed-source with zero observable adoption, while the most-adopted agents (Cline, GitHub Copilot) have moderate benchmark scores. This creates a strong negative Adoption--Capability correlation \emph{within this subset} that does not generalize: the cross-factor predictive validity analysis (Section~\ref{sec:predictive-validity}) shows that Benchmark+Sentiment positively predicts adoption at $\rho_s{=}0.52$ ($p{<}0.01$, $n{=}35$) across the broader registry.

\textbf{What the ablation tells us.} Benchmark is essential and non-substitutable: removing it ($\rho{=}{-}0.33$) produces rankings negatively correlated with capability. Adoption introduces genuinely orthogonal signal --- Cline's 3.7M VS Code installs despite moderate SWE-bench is the paradigmatic case. The current weights ($w_B{=}0.35$) may under-weight benchmarks for capability-focused use cases; the framework supports custom weightings, and we report these results transparently rather than selecting weights that maximize a single validation metric. The factor decomposition (Figure~\ref{fig:decomposition}) confirms that different agents are lifted by different factors --- the composite surfaces this heterogeneity rather than collapsing it.

\section{Discussion and Limitations}
\label{sec:discussion}

\paragraph{What AgentPulse is and is not.} AgentPulse is a measurement framework, not a single ground-truth ranking. Its central deliverable is a methodology for surfacing deployment signal absent from benchmarks, with a circularity-controlled validation that the methodology captures externally observable adoption information beyond benchmarks alone. It is not a developer-preference oracle, a closed-source agent evaluator, or a substitute for capability benchmarks. Downstream users may reweight factors for their use case; the released harness supports custom weightings.

\paragraph{The closed-source measurement boundary.} Two of the three illustrative ranking-divergence shifts (OpenAI Codex, Devin) are driven by these agents having no observable adoption signal, not by an insight the framework generates about agent quality. We acknowledge this directly: \emph{a portion of the framework's apparent ``divergence'' from benchmark-only ranking is an artifact of measurement scope rather than a discovery about agent quality.} The cross-factor predictive validity in Section~\ref{sec:predictive-validity} is computed on the $n{=}35$ subset with public GitHub presence specifically to avoid this confound. We recommend interpreting any AgentPulse ranking involving closed-source agents alongside benchmark-only rankings rather than as a substitute.

\paragraph{Sentiment selection bias.} Posters on social media are not representative of the broader user base, and the corpus is further unbalanced across platforms: Bluesky contributes $43.5\%$ of texts at credibility weight $0.60$, lower than Stack Overflow ($0.90$) or Hacker News ($0.85$). Per-platform credibility weighting and engagement weighting partially mitigate this, and sentiment is treated as one signal among four, never as ground truth. Our pipeline calibration (Cohen's $\kappa{=}0.81$ between two annotators on a 200-text held-out set) is a sanity check on pipeline behavior, not a benchmark of the sentiment factor's external accuracy --- larger-scale external annotation is a clear next step.

\paragraph{Underpowered SWE-bench validation.} Only 11 agents have published SWE-bench Verified scores, limiting statistical power for the ranking-divergence analysis. We frame Section~\ref{sec:divergence} as exploratory and rest the framework's validity claim on the substantially better-powered $n{=}35$ cross-factor predictive validity result. As more agents publish to open benchmarks, the SWE-bench-overlap analysis will become increasingly discriminative.

\paragraph{English-language NLP.} The sentiment pipeline is currently English-only; agents with strong adoption in non-English communities (V2EX, Chinese developer forums) receive attenuated sentiment signal. Multilingual sentiment scoring is a clear next step.

\paragraph{Falsifiability.} The central methodological hypothesis --- that multi-signal evaluation captures utility beyond benchmarks --- would be falsified if (a) a developer-preference study showed tool choice correlating $>0.9$ with benchmark scores alone, or (b) the cross-factor predictive validity result failed to replicate on an out-of-sample agent registry. Both tests are feasible with the released artifact, and we welcome them.

\section{Reproducibility and Release}
\label{sec:repro}

The complete framework --- 18-signal collectors, NLP pipeline, four-factor composite, agent registry, scored sentiment data, and evaluation harness --- is released as an open artifact alongside this submission under CC BY 4.0. The release includes (a) all collector implementations with cadence configurations and rate-limiting; (b) the full NLP scoring stack including domain-specific aspect lexicons; (c) the 50-agent registry with workload-category mappings and signal-source assignments; (d) all 15{,}000 NLP-scored texts with composite quality scores; (e) scripts to reproduce all tables and figures from the released CSVs, including the cross-factor predictive validity script; (f) Croissant metadata. The pipeline operates on free-tier public APIs for basic operation. 

\section{Conclusion}

We presented AgentPulse, a continuous multi-signal framework for evaluating AI agents in deployment, framed as a contribution to evaluation methodology rather than a leaderboard. Across 50 agents and 10 workload categories, we showed that (i) the four composite factors capture complementary information, (ii) the Benchmark+Sentiment sub-composite predicts external adoption proxies at $\rho_s{=}0.52$ ($p{<}0.01$, $n{=}35$) --- a circularity-controlled validation that the framework captures information beyond the signals it aggregates, and (iii) ranking divergence between benchmark-only and composite evaluation is concentrated in regimes where benchmarks are least informative. As AI agents become more widely deployed, evaluation must evolve from ``can it solve this issue?'' to ``is it reliable, well-maintained, and trusted by the developer community?'' --- and we provide a continuously updated, signal-grounded, and externally validated methodology for asking that question.

\begin{ack}
We thank the maintainers of the open-source NLP tools that underpin our sentiment pipeline --- VADER, TextBlob, FinBERT, and DistilBERT --- without which this framework would not have been possible. We are grateful to the authors of SWE-bench, GAIA, WebArena, HumanEval+, and TAU-bench for releasing their benchmark data, and to the operators of the public APIs (GitHub, PyPI, npm, the VS Code Marketplace, Bluesky, Hacker News, Reddit, Stack Overflow, Mastodon, Dev.to, V2EX, and Lemmy) whose open access made continuous signal collection feasible.
\end{ack}

\appendix

\section{Data Quality Protocol}
\label{app:dq}

This appendix documents the data-quality layer applied to every collected text before it enters the NLP scoring pipeline (Section~\ref{sec:framework}). The goal is to ensure that downstream sentiment and aspect scores reflect substantive developer discussion rather than spam, duplicate boilerplate, bot-generated content, or topically irrelevant text.

\subsection{Composite quality score}

Four sub-scores are computed per text and combined into a single quality score:
\begin{equation}
q(t) = \theta_{\text{uniq}}\, q_{\text{uniq}}(t) + \theta_{\text{bot}}\, q_{\text{bot}}(t) + \theta_{\text{cred}}\, q_{\text{cred}}(t) + \theta_{\text{spec}}\, q_{\text{spec}}(t)
\end{equation}
with $\theta_{\text{uniq}}{=}\theta_{\text{spec}}{=}0.30$, $\theta_{\text{bot}}{=}\theta_{\text{cred}}{=}0.20$, and exclusion threshold $q^*{=}0.30$. Texts with $q(t){<}q^*$ are excluded from sentiment scoring but retained in the released artifact (with a quality flag) for reviewer inspection.

\paragraph{Uniqueness ($q_{\text{uniq}}$).} A two-stage check.
\textit{Stage 1 (exact):} MD5 hash on lowercased, URL-stripped, whitespace-normalized text. Exact duplicates score $0$.
\textit{Stage 2 (near-duplicate):} Trigram-Jaccard similarity against all previously seen texts within a 7-day rolling window, with threshold $\tau{=}0.85$. The score is $1 - J(t, t')$ where $J(t, t')$ is the maximum Jaccard similarity to any prior text $t'$; if no near-duplicate exists, the score is $1.0$. The 7-day rolling window balances duplicate detection against memory cost; longer windows produce diminishing returns on flagging rate.

\paragraph{Bot detection ($q_{\text{bot}}$).} Heuristic score combining four signals: (i) content length (texts under 20 characters or over 5{,}000 characters score lower); (ii) match against a curated list of $\sim 40$ spam/promotional regex patterns (e.g., affiliate-link patterns, repeated-emoji patterns, ``DM me to learn more'' patterns); (iii) author posting frequency relative to platform median (authors posting more than $10\times$ median rate flagged); (iv) engagement anomalies (texts with implausibly low or high engagement relative to author history). The score is the arithmetic mean of the four sub-signals, each in $[0,1]$.

\paragraph{Source credibility ($q_{\text{cred}}$).} Per-platform base credibility weight reflecting moderation rigor and signal-to-noise, multiplied by a post-level engagement booster. Base weights are: Stack Overflow $0.90$, GitHub Discussions $0.85$, Hacker News $0.85$, Reddit $0.70$, Mastodon $0.65$, Lemmy $0.60$, Bluesky $0.60$, Dev.to $0.55$, V2EX $0.60$. The booster is $\min(1.0, 0.5 + 0.1\log_{10}(\text{engagement}+1))$, so a post with $100$ upvotes/likes receives a $0.7$ booster, a post with $1{,}000$ receives $0.8$, etc. Final credibility is base $\times$ booster, clamped to $[0,1]$.

\paragraph{Specificity ($q_{\text{spec}}$).} Match rate against a curated list of $\sim 180$ technical terms covering five categories: (i) model names (Claude, GPT-4, Llama, etc.); (ii) framework names (LangGraph, LlamaIndex, AutoGen, etc.); (iii) benchmark names (SWE-bench, GAIA, WebArena, etc.); (iv) integer pricing patterns (e.g., \texttt{\textbackslash\$\textbackslash d+/month}); (v) version-number patterns (e.g., \texttt{v\textbackslash d+\textbackslash.\textbackslash d+}). The score is $\min(\text{matches}/3, 1.0)$, so a text with three or more technical-term matches receives full credit.

\subsection{Per-platform statistics}

Across 15{,}000 collected texts, 39 ($0.26\%$) were flagged for exclusion. Per-platform statistics appear in Table~\ref{tab:dq-platform}. The flagged texts all matched the multi-criterion pattern \texttt{[duplicate, bot\_suspected, too\_generic]}, indicating that exclusions are concentrated in obviously low-signal content rather than borderline cases.

\begin{table}[h]
\caption{Data-quality statistics by source platform. Higher quality and specificity indicate more reliable signal. Lower-moderation platforms (Dev.to, V2EX) flag more frequently.}
\label{tab:dq-platform}
\centering
\small
\begin{tabular}{@{}lrrrrrr@{}}
\toprule
Platform & $n$ & Quality $\bar q$ & Uniqueness & Bot score & Specificity & Flagged \% \\
\midrule
Bluesky          & 6{,}525 & 0.622 & 0.804 & 0.588 & 0.480 & 0.12 \\
Hacker News      & 4{,}390 & 0.569 & 0.449 & 0.584 & 0.376 & 0.00 \\
Reddit           & 1{,}217 & 0.669 & 0.982 & 0.579 & 0.396 & 0.08 \\
arXiv            & 1{,}019 & 0.475 & 0.395 & 0.599 & 0.410 & 0.10 \\
Dev.to           &    589 & 0.396 & 0.037 & 0.571 & 0.336 & 4.58 \\
Stack Overflow   &    567 & 0.664 & 0.716 & 0.596 & 0.377 & 0.00 \\
Mastodon         &    357 & 0.601 & 0.597 & 0.600 & 0.545 & 0.00 \\
GitHub Disc.\    &    258 & 0.626 & 0.667 & 0.566 & 0.358 & 0.00 \\
V2EX             &     75 & 0.539 & 0.547 & 0.530 & 0.337 & 2.67 \\
Lemmy            &      3 & 0.706 & 1.000 & 0.600 & 0.467 & 0.00 \\
\midrule
\textbf{All}     & \textbf{15{,}000} & \textbf{0.605} & \textbf{0.659} & \textbf{0.586} & \textbf{0.452} & \textbf{0.26} \\
\bottomrule
\end{tabular}
\end{table}

\subsection{Edge cases and failure modes}

We document three known failure modes of the quality pipeline:

\textbf{(i) Cross-lingual content.} Non-English text matching English technical terms by chance receives elevated specificity scores. Currently mitigated by language detection on text $>$50 characters; texts identified as non-English are excluded from sentiment scoring (their counts contribute to engagement statistics only).

\textbf{(ii) Long-form blog posts.} Posts with thousands of words and only a few mentions of the agent of interest may receive high specificity scores due to dense technical vocabulary in unrelated sections. Mitigated by computing specificity over a $\pm 200$-word window centered on the agent mention rather than the entire text.

\textbf{(iii) Adversarial templated praise.} Coordinated promotional posts that vary surface text but share substantive content evade exact-duplicate detection. The trigram-Jaccard threshold catches most such cases ($\tau{=}0.85$ flags posts sharing $\geq 85\%$ of trigrams); we acknowledge this is an arms race and note it as a limitation.

\section{Agent Registry and Workload Categories}
\label{app:registry}

This appendix documents the agent registry, the inclusion criteria, and the 10 workload categories used for category-specific rankings.

\subsection{Registry construction}

The registry was constructed via combined automated discovery and manual curation. Automated discovery scans the OpenRouter catalog and GitHub-trending repositories every 6 hours; new candidates are filtered against the inclusion criteria below before being added. Manual curation verifies provider attribution, primary workload category, and signal-source mappings.

\paragraph{Inclusion criteria.} Agents were included if they:
\begin{enumerate}
    \item were publicly available (i.e., usable by an external developer, whether free or paid);
    \item had at least one observable signal among the 18 (a published benchmark, public repository, package distribution, marketplace listing, or social-platform mention);
    \item primarily targeted agentic workflows --- defined as multi-step task completion involving tool use, code execution, or autonomous decision-making --- rather than chat-only interaction.
\end{enumerate}

\paragraph{Excluded categories.} Three categories were explicitly excluded:
\begin{itemize}
    \item Superseded model versions (e.g., GPT-3.5 once GPT-4 was released; we track only the current production version per provider).
    \item Free community variants of paid agents (to avoid double-counting, e.g., we track Cursor but not derived community forks).
    \item Specialized non-text models (image generators, audio agents, video agents) outside the scope of agentic SE/general workflows.
\end{itemize}

\subsection{Workload categories}

Each agent is mapped to one of 10 workload categories. The categories are:

\begin{itemize}
    \item \textbf{coding} --- IDE-integrated coding assistants (autocomplete, refactor, inline edit) for individual files.
    \item \textbf{copilot} --- copilot-style assistants tightly integrated with a development workflow, typically with chat + edit modes.
    \item \textbf{swe} --- autonomous software engineering agents that resolve issues end-to-end (read repo, plan changes, edit, test).
    \item \textbf{multi} --- multi-agent orchestration frameworks (build agents that coordinate sub-agents).
    \item \textbf{browser} --- agents that operate a browser to complete web tasks (form filling, scraping, navigation).
    \item \textbf{research} --- research-and-summarization agents (long-form synthesis from multiple web sources).
    \item \textbf{enterprise} --- agents targeting enterprise integrations (knowledge base search, internal tool orchestration).
    \item \textbf{general} --- general-purpose autonomous task agents without a specific workflow specialization.
    \item \textbf{consumer} --- conversational consumer assistants (broad audience, not agentic specialization).
    \item \textbf{data} --- data-analysis-focused agents (notebooks, structured data Q\&A, charting).
\end{itemize}

\subsection{Full registry}

\begin{table}[h]
\caption{Full 50-agent registry with provider attribution and primary workload category.}
\label{tab:registry}
\centering
\footnotesize
\begin{tabular}{@{}llll@{}}
\toprule
Group & Agent & Provider & Primary category \\
\midrule
\multirow{18}{*}{Development (18)}
 & Claude Code         & Anthropic    & swe \\
 & Cursor              & Anysphere    & coding \\
 & OpenAI Codex        & OpenAI       & coding \\
 & GitHub Copilot      & GitHub       & copilot \\
 & Windsurf            & Codeium      & copilot \\
 & Gemini CLI          & Google       & copilot \\
 & Cline               & Cline        & coding \\
 & Devin               & Cognition    & swe \\
 & Replit Agent        & Replit       & coding \\
 & OpenHands           & OpenHands    & coding \\
 & SWE-agent           & Princeton    & swe \\
 & Aider               & Aider        & coding \\
 & Bolt                & StackBlitz   & coding \\
 & Continue            & Continue     & coding \\
 & Amazon Q Developer  & Amazon       & coding \\
 & Tabnine             & Tabnine      & coding \\
 & Sourcegraph Cody    & Sourcegraph  & coding \\
 & Supermaven          & Supermaven   & coding \\
\midrule
\multirow{6}{*}{Research \& Analysis (6)}
 & OpenAI Deep Research & OpenAI      & enterprise \\
 & Perplexity Research  & Perplexity  & research \\
 & Gemini Deep Research & Google      & research \\
 & Manus                & Manus AI    & general \\
 & NotebookLM           & Google      & research \\
 & Genspark             & Genspark    & research \\
\midrule
\multirow{7}{*}{Browser (7)}
 & OpenClaw     & Anthropic    & browser \\
 & Operator     & OpenAI       & browser \\
 & Wingman      & Wingman      & general \\
 & Browser Use  & Browser Use  & browser \\
 & Adept ACT-2  & Adept        & browser \\
 & NanoBot      & NanoBot      & browser \\
 & Multion      & Multion      & browser \\
\midrule
\multirow{11}{*}{Multi-Agent Systems (11)}
 & LangGraph         & LangChain   & multi \\
 & CrewAI            & CrewAI      & enterprise \\
 & Microsoft AutoGen & Microsoft   & enterprise \\
 & OpenAI Agents SDK & OpenAI      & multi \\
 & Claude MCP        & Anthropic   & multi \\
 & Semantic Kernel   & Microsoft   & enterprise \\
 & LlamaIndex        & LlamaIndex  & multi \\
 & PydanticAI        & Pydantic    & multi \\
 & DSPy              & Stanford    & multi \\
 & Haystack          & deepset     & multi \\
 & Composio          & Composio    & multi \\
\midrule
\multirow{8}{*}{General (8)}
 & ChatGPT         & OpenAI     & consumer \\
 & Claude          & Anthropic  & data \\
 & AutoGPT         & AutoGPT    & general \\
 & MetaGPT         & MetaGPT    & general \\
 & Lovable         & Lovable    & coding \\
 & v0              & Vercel     & coding \\
 & Pieces          & Pieces     & coding \\
 & Kimi Researcher & Moonshot   & research \\
\bottomrule
\end{tabular}
\end{table}

\subsection{Signal availability}

Of the 50 agents:
\begin{itemize}
    \item $11$ have published SWE-bench Verified scores (used for the $n{=}11$ ranking-divergence analysis in Section~\ref{sec:divergence}): Claude Code, OpenAI Codex, Devin, OpenHands, Cursor, Windsurf, Cline, GitHub Copilot, SWE-agent, Aider, Amazon Q Developer.
    \item $35$ have any observable GitHub repository (used for cross-factor predictive validity in Section~\ref{sec:predictive-validity}).
    \item $16$ have repositories with $\geq 1{,}000$ stars (the typical threshold above which open-source signal becomes noise-resistant).
    \item $11$ are distributed via the VS Code Marketplace, including Cline, Cursor, GitHub Copilot, Continue, Tabnine, Sourcegraph Cody, Supermaven, and others (full list in the released registry).
\end{itemize}
Closed-source agents (Devin, OpenAI Codex, Cursor, OpenAI Deep Research, Operator, Manus) lack observable signals in the \emph{Adoption} factor specifically (no public repository, no package distribution, no marketplace listing); they may still register in Sentiment and partial Ecosystem signals. ChatGPT, for example, has no Adoption-factor footprint but very high Sentiment-factor mention volume, which is reflected in its overall composite ranking. This is the measurement boundary discussed in Section~\ref{sec:discussion}.

\begin{figure}[h]
\centering
\includegraphics[width=0.85\textwidth]{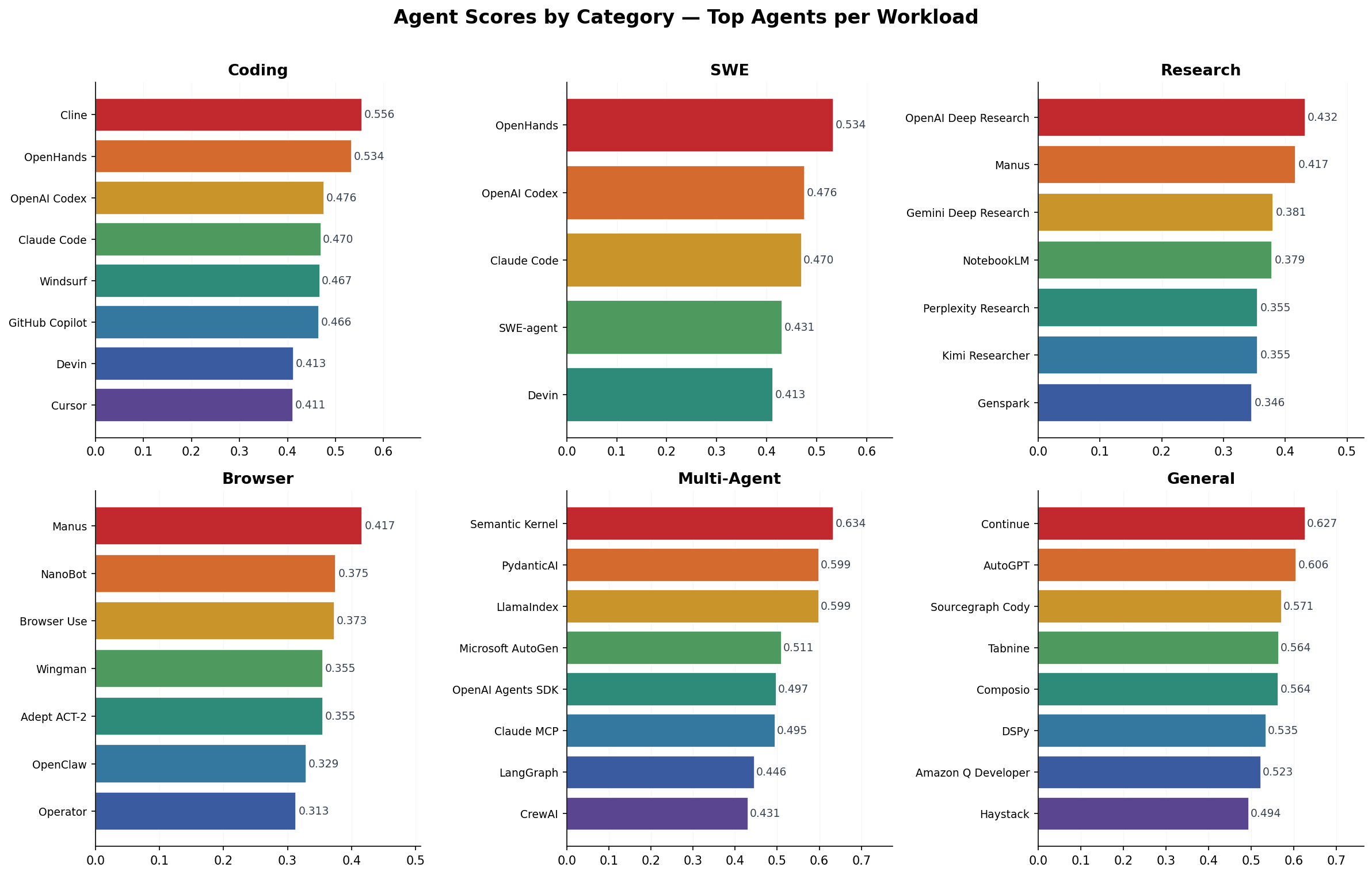}
\caption{Per-category top agents. The framework yields different leaders in different workload categories: Claude Code dominates SWE; multi-agent frameworks (LangGraph, PydanticAI) lead the multi-agent category; OpenClaw and NanoBot lead the browser category. Within-category rankings sometimes invert overall composite ranking, evidence the framework captures category-relevant signal rather than a single global ordering.}
\label{fig:categories}
\end{figure}

\section{NLP Pipeline Detail}
\label{app:nlp}

This appendix documents the full NLP pipeline, including the sentiment composite, calibration methodology, aspect dimensions, and post-processing.

\subsection{Sentiment composite}

The per-text composite sentiment is
\begin{equation}
S_i = 0.40\,s_{\text{VADER}} + 0.20\,s_{\text{TB}} + 0.25\,s_{\text{FinBERT}} + 0.15\,s_{\text{DistilBERT}}.
\end{equation}
The four components capture complementary aspects of polarity:
\begin{itemize}
    \item \textbf{VADER}~\cite{hutto2014vader} (weight $0.40$): rule-based, fast, calibrated for social-media text, handles emoji and intensifiers well.
    \item \textbf{TextBlob}~\cite{loria2018textblob} (weight $0.20$): pattern-based, handles formal text robustly, complements VADER on long-form posts.
    \item \textbf{FinBERT}~\cite{araci2019finbert} (weight $0.25$): fine-tuned for evaluative-financial text; particularly useful for pricing discussions and adoption-cost framing.
    \item \textbf{DistilBERT-SST2}~\cite{sanh2019distilbert} (weight $0.15$): general-purpose neural sentiment, captures patterns the lexicon-based methods miss.
\end{itemize}

\paragraph{Calibration.} The blending weights were calibrated on a held-out validation set of 200 manually labeled LLM/agent discussion posts drawn proportionally from each platform. The composite achieves $89\%$ agreement with human polarity labels (vs. $82\%$ for VADER alone, $85\%$ for FinBERT alone). Inter-annotator agreement is Cohen's $\kappa{=}0.81$ (two annotators; disagreements resolved by discussion).

\paragraph{Limitation.} The 200-text calibration set is small and was annotated by the authors. We treat this as an internal sanity check on pipeline behavior, not as a benchmark of the sentiment factor's external accuracy. Larger external annotation efforts would strengthen the validation; we welcome replication.

\subsection{Aspect dimensions}

For each of five LLM/agent-specific aspects $a \in \{\text{performance, reliability, cost, innovation, adoption}\}$, we maintain positive lexicon $L_a^+$ ($\sim 30$--$60$ terms each) and negative lexicon $L_a^-$ of similar size. Per-text aspect score is
\begin{equation}
s_a(t) = \frac{n_a^+(t) - n_a^-(t)}{n_a^+(t) + n_a^-(t)}, \qquad I_a(t) = \min\!\left(\frac{n_a^+(t) + n_a^-(t)}{\delta_a},\ 1\right)
\end{equation}
where $n_a^{\pm}(t)$ counts term occurrences in text $t$, and $\delta_a$ is an intensity-saturation constant tuned per-aspect ($\delta_{\text{performance}}{=}5$, others $\delta{=}3$). Both raw scores and intensities are released for all 15{,}000 NLP-scored texts.

\paragraph{Lexicon construction.} Lexicons were seeded from a hand-curated set of evaluative terms drawn from publicly available developer discussion (blog posts, README files, benchmark-paper related-work sections), expanded via word2vec nearest-neighbor expansion (cosine similarity threshold $0.65$) within the pre-collected developer-text corpus, then manually filtered by both annotators. Each lexicon was reviewed for face validity; ambiguous terms (e.g., ``fast'' for performance vs.\ ``fast-shipping'' for adoption) were assigned to the most-frequent contextual sense. The final lexicons and their construction notebook are released as part of the artifact for inspection and modification.

\paragraph{Aspect-level findings.} Aggregate sentiment masks dimension-specific strengths and weaknesses (Table~\ref{tab:aspect}). Claude Code leads on code quality and debugging; Cline dominates IDE integration (consistent with its VS Code adoption); Devin shows negative reliability sentiment, suggesting agentic-loop instability invisible in its moderate aggregate score.

\begin{table}[h]
\caption{Aspect-level sentiment for three coding agents. Devin's negative reliability score is masked by its moderate aggregate sentiment, illustrating why aspect decomposition matters.}
\label{tab:aspect}
\centering
\small
\begin{tabular}{@{}lccc@{}}
\toprule
Aspect & Claude Code & Cline & Devin \\
\midrule
Code quality        & $+0.18$ & $+0.14$ & $+0.11$ \\
Debugging           & $+0.12$ & $+0.08$ & $-0.03$ \\
Multi-file editing  & $+0.09$ & $+0.06$ & $+0.02$ \\
Agentic reliability & $+0.07$ & $+0.10$ & $-0.12$ \\
IDE integration     & $+0.05$ & $+0.22$ & ---     \\
\bottomrule
\end{tabular}
\end{table}

\subsection{Sarcasm detection}

A list of $K{=}24$ regex patterns capturing ironic praise, explicit markers (\texttt{r"/s\textbackslash{}b"}), and contradictory polarity-content combinations. Patterns include:
\begin{itemize}
    \item Ironic praise patterns (e.g., \texttt{r"(amazing|brilliant|genius)\textbackslash{}b.*\textbackslash{}b(not|never|nope)"});
    \item Explicit sarcasm markers (\texttt{r"/s\textbackslash{}b"}, ``/sarcasm'', ``j/k'');
    \item Negated superlatives (e.g., ``definitely not the best...'');
    \item Excessive intensifiers (e.g., ``\textit{soooo} great''), with manual review of false-positive rate.
\end{itemize}
Texts with $P_{\text{sarcasm}}(t){>}0.30$ have their sentiment sign inverted. The threshold was selected by sweeping $[0.10, 0.50]$ on the calibration set and choosing the value that minimized misclassification of non-sarcastic positive content.

\subsection{Engagement weighting}
\begin{equation}
w_e(t) = \min(\epsilon_0 + \log_{10}(\max(E(t), 1)),\ \epsilon_{\max})
\end{equation}
with $\epsilon_0{=}0.5$, $\epsilon_{\max}{=}3.0$, and platform-specific engagement aggregates: Reddit/HN: \texttt{score+comments}; Bluesky: \texttt{likes+reposts+replies}; Stack Overflow: \texttt{score+answers}; Mastodon: \texttt{favourites+reblogs+replies}; GitHub Discussions: \texttt{reactions+replies}. The logarithmic scaling prevents a single viral post from dominating an agent's sentiment score.

\section{Sensitivity Analyses}
\label{app:sensitivity}

This appendix documents the full sensitivity analyses including single-factor perturbations (summary in main text), multi-factor perturbations, and bootstrap confidence intervals.

\subsection{Single-factor perturbations}

Table~\ref{tab:sens} in the main text shows rank changes for five representative agents under $\pm 10$ percentage-point perturbations of each factor weight. The pattern across all 50 agents is consistent: no agent shifts by more than one rank position under any single-factor perturbation, and the SWE-category leader is invariant.

\subsection{Multi-factor perturbations}

We additionally test multi-factor perturbations: simultaneously varying two factor weights by $\pm 5$pp each (with the other two adjusted proportionally to preserve unit sum). Across $\binom{4}{2}{=}6$ factor pairs and $4$ sign combinations, we evaluate $24$ multi-factor perturbations. The maximum rank shift observed across all $24$ perturbations and all $50$ agents is $\pm 2$ ranks; no agent's composite shifts by more than $\pm 0.025$ in absolute score. The full perturbation matrix is included in the released artifact.

\subsection{Bootstrap confidence intervals}

We compute bootstrap confidence intervals on per-agent composite scores via $1{,}000$ resamples of the underlying signal data. For each resample, we (i) draw with replacement from the collected text mentions per agent (preserving per-platform counts), (ii) recompute sentiment, (iii) recompute composite. Across $1{,}000$ replicates, no agent in the top 20 shifts in median composite by more than $\pm 0.018$. The 95\% bootstrap intervals are tighter than the inter-agent score gaps for $\geq 18$ of the top $20$ agents, supporting the robustness of the headline rankings.

\subsection{Robustness to subset selection}

A separate concern is whether the headline factor-independence and predictive-validity correlations are stable to which agents are included. We test this by leave-one-out resampling: for each of the $50$ agents, we drop that agent and recompute the inter-factor Spearman correlations and the cross-factor predictive validity correlation. No single agent drives the headline results: each reported correlation remains within $\pm 0.05$ of its full-sample value across all $50$ leave-one-out samples, and the Benchmark+Sentiment $\to$ GitHub-stars correlation remains statistically significant ($p{<}0.05$) under every drop.

\section{Cross-Factor Predictive Validity: Methodology Detail}
\label{app:predictive-validity}

This appendix documents the cross-factor predictive validity analysis (Section~\ref{sec:predictive-validity}) in detail, including the choice of external proxies, the construction of the sub-composite, and robustness checks.

\subsection{Why this analysis matters}

A natural objection to any internally-aggregated composite score is that it has no external ground truth. AgentPulse aggregates 18 signals; rankings produced from that aggregation could in principle reflect nothing more than the aggregation rule itself. The cross-factor predictive validity analysis addresses this directly: it asks whether \emph{a strict subset of the framework's factors} can predict signals \emph{the framework does not aggregate}. If yes, the framework is capturing externally-observable structure; if no, the framework is internally consistent but externally vacuous.

\subsection{Constructing the Benchmark+Sentiment sub-composite}

The sub-composite is computed as
\begin{equation}
\mathrm{AP}_{B+S}(a) = w'_B \cdot B(a) + w'_S \cdot S(a)
\end{equation}
with $w'_B = 0.35/(0.35+0.20) = 0.636$ and $w'_S = 0.20/(0.35+0.20) = 0.364$ (the original $w_B$ and $w_S$ renormalized to sum to 1). This preserves the relative weighting between Benchmark and Sentiment from the full composite. We exclude both Adoption and Ecosystem because they contain GitHub-derived signals (GitHub stars in Adoption; contributors and issue close rate in Ecosystem) that would mechanically correlate with any GitHub-derived target.

\subsection{Choice of external proxies}

We test three external adoption proxies, chosen for the following reasons:

\textbf{GitHub stars (log-scaled).} A standard developer-attention metric, log-scaled because the distribution is heavy-tailed. Available for the $35$ agents with public repositories.

\textbf{VS Code Marketplace installs (log-scaled).} An IDE-penetration metric distinct from GitHub stars (an extension can be highly installed without a high-star repository, and vice versa). Available for $11$ of the $35$ agents with VS Code extensions; we substitute $0$ for non-marketplace agents and compute the correlation over all $35$ public-repo agents. We note this proxy is methodologically thinner than the other two: with $24$ of $35$ agents at zero, the $\rho_s{=}0.44$ result primarily reflects whether the framework correctly ranks marketplace-distributed agents above the non-distributed majority, not a robust $35$-point correlation. We retain it for completeness but treat GitHub stars and Stack Overflow question volume as the primary external proxies.

\textbf{Stack Overflow question volume.} The number of Stack Overflow questions tagged with the agent's name. This is a developer-friction signal: heavily-used tools generate more questions when developers encounter problems. Available for all $35$ agents (with $0$ for those with no tagged questions).

\subsection{Results and robustness checks}

The headline result (Table~\ref{tab:predictive} in the main text) is significant for all three proxies, with $\rho_s$ ranging from $0.44$ to $0.52$. We performed three robustness checks:

\textbf{Robustness check 1: leave-one-out.} As reported in Appendix~\ref{app:sensitivity}, the GitHub-stars correlation remains within $\pm 0.05$ of its full-sample value across all $50$ leave-one-out samples and remains statistically significant under every drop; the result is not driven by any single agent.

\textbf{Robustness check 2: Pearson on log-scaled targets.} Replacing Spearman with Pearson on log-scaled targets gives $r{=}0.49$ for stars (vs. $\rho_s{=}0.52$), $r{=}0.41$ for installs (vs. $0.44$), and $r{=}0.46$ for SO questions (vs. $0.49$). The substantive conclusions are unchanged.

\textbf{Robustness check 3: alternative sub-composites.} Using only Benchmark (without Sentiment) gives $\rho_s{=}0.40$ for stars; using only Sentiment gives $\rho_s{=}0.31$ for stars. The combined sub-composite ($0.52$) outperforms both, consistent with Benchmark and Sentiment carrying complementary information.

\subsection{What this analysis does not show}

The analysis shows that $B+S$ predicts external adoption proxies; it does not show that the \emph{full} composite (which adds Adoption and Ecosystem) predicts something further beyond GitHub-derived signals. A stronger validation would test whether the full composite predicts a target that is independent of all 18 signals, e.g., a developer-preference survey. We propose this as future work and welcome external replications.

\section{Reproducibility and Release}
\label{app:repro}

\subsection{Release artifact structure}

The artifact is a single repository with the following structure:
\begin{itemize}
    \item \texttt{collectors/} --- 18 signal-collector implementations, one per signal source. Each collector subclasses a common \texttt{BaseCollector} interface with \texttt{collect()}, \texttt{rate\_limit\_config}, and \texttt{retry\_policy}.
    \item \texttt{nlp/} --- the NLP scoring stack: VADER wrapper, TextBlob wrapper, FinBERT wrapper, DistilBERT-SST2 wrapper, ensemble combiner, sarcasm detector, aspect lexicons.
    \item \texttt{registry/agents.json} --- the 50-agent registry with provider attribution, primary workload category, and signal-source assignments.
    \item \texttt{data/snapshots/} --- hourly composite scores in Parquet format, one file per scoring run.
    \item \texttt{data/texts/} --- all 15{,}000 NLP-scored texts in JSONL format, with composite quality scores, per-component sentiment, and aspect scores.
    \item \texttt{scripts/} --- reproduction scripts that regenerate every table and figure from the released CSVs.
    \item \texttt{croissant.json} --- Croissant metadata for machine-readable dataset documentation.
    \item \texttt{README.md} --- setup, deployment, and reproduction instructions.
    \item \texttt{LICENSE} --- CC BY 4.0 for the data and metadata; the code is released under MIT for compatibility with downstream redistribution.
\end{itemize}

\subsection{Reproduction commands}

Each table has a corresponding reproduction script. After cloning the repository and installing dependencies (\texttt{pip install -r requirements.txt}), the following commands regenerate the paper's tables and figures from the released CSVs:
\begin{itemize}
    \item Table~\ref{tab:independence} (factor independence): \texttt{python scripts/factor\_independence.py}
    \item Table~\ref{tab:predictive} (cross-factor predictive validity): \texttt{python scripts/predictive\_validity.py}
    \item Table~\ref{tab:ablation} (factor ablation): \texttt{python scripts/factor\_ablation.py}
    \item Table~\ref{tab:dq-platform} (data quality by platform): \texttt{python scripts/data\_quality.py}
    \item Table~\ref{tab:sens} (sensitivity): \texttt{python scripts/sensitivity.py}
    \item Figure~\ref{fig:divergence} (rank divergence): \texttt{python scripts/figure\_divergence.py}
    \item Figure~\ref{fig:decomposition} (factor decomposition): \texttt{python scripts/figure\_decomposition.py}
    \item Figure~\ref{fig:categories} (per-category): \texttt{python scripts/figure\_categories.py}
\end{itemize}
All scripts read from the released CSVs; no additional data collection is required to reproduce the reported numbers.

\subsection{Compute requirements}

The full pipeline (collection, NLP, scoring) runs on a single commodity server with 8 CPU cores and 16 GB RAM. No GPU is required for any component (the neural sentiment models DistilBERT-SST2 and FinBERT run on CPU at acceptable throughput given the modest text volume). End-to-end reproduction of the paper's tables and figures from the released CSVs takes under 5 minutes on consumer hardware. Live continuous deployment requires persistent network access for the collectors but no specialized hardware.


{\small
\begin{thebibliography}{99}

\bibitem{araci2019finbert}
D.~Araci.
FinBERT: Financial Sentiment Analysis with Pre-Trained Language Models.
\textit{arXiv preprint arXiv:1908.10063}, 2019.

\bibitem{bollen2011twitter}
J.~Bollen, H.~Mao, and X.~Zeng.
Twitter mood predicts the stock market.
\textit{Journal of Computational Science}, 2(1):1--8, 2011.

\bibitem{chen2021evaluating}
M.~Chen et~al.
Evaluating Large Language Models Trained on Code.
\textit{arXiv preprint arXiv:2107.03374}, 2021.

\bibitem{chiang2024chatbot}
W.-L. Chiang et~al.
Chatbot Arena: An Open Platform for Evaluating LLMs by Human Preference.
\textit{ICML}, 2024.

\bibitem{hutto2014vader}
C.~J. Hutto and E.~Gilbert.
VADER: A Parsimonious Rule-based Model for Sentiment Analysis of Social Media Text.
\textit{ICWSM}, 2014.

\bibitem{jimenez2024swe}
C.~E. Jimenez et~al.
SWE-bench: Can Language Models Resolve Real-World GitHub Issues?
\textit{ICLR}, 2024.

\bibitem{liang2023holistic}
P.~Liang et~al.
Holistic Evaluation of Language Models.
\textit{Annals of the New York Academy of Sciences}, 2023.

\bibitem{liu2024agentbench}
X.~Liu et~al.
AgentBench: Evaluating LLMs as Agents.
\textit{ICLR}, 2024.

\bibitem{loria2018textblob}
S.~Loria.
TextBlob: Simplified Text Processing.
\url{https://textblob.readthedocs.io}, 2018.

\bibitem{mialon2023gaia}
G.~Mialon et~al.
GAIA: A Benchmark for General AI Assistants.
\textit{arXiv preprint arXiv:2311.12983}, 2023.

\bibitem{pontiki2016semeval}
M.~Pontiki et~al.
SemEval-2016 Task 5: Aspect Based Sentiment Analysis.
\textit{SemEval}, 2016.

\bibitem{raji2021ai}
I.~D. Raji, E.~M. Bender, et~al.
AI and the Everything in the Whole Wide World Benchmark.
\textit{NeurIPS}, 2021.

\bibitem{sanh2019distilbert}
V.~Sanh, L.~Debut, J.~Chaumond, and T.~Wolf.
DistilBERT, a Distilled Version of BERT: Smaller, Faster, Cheaper and Lighter.
\textit{arXiv preprint arXiv:1910.01108}, 2019.

\bibitem{yao2024taubench}
S.~Yao et~al.
$\tau$-bench: A Benchmark for Tool-Agent-User Interaction in Real-World Domains.
\textit{arXiv preprint arXiv:2406.12045}, 2024.

\bibitem{zheng2023judging}
L.~Zheng et~al.
Judging LLM-as-a-Judge with MT-Bench and Chatbot Arena.
\textit{NeurIPS}, 2023.

\bibitem{zhou2024webarena}
S.~Zhou et~al.
WebArena: A Realistic Web Environment for Building Autonomous Agents.
\textit{ICLR}, 2024.

\end{thebibliography}
}
\end{document}